%% file: acl2020.tex
\documentclass[11pt,a4paper]{article}
\usepackage{authblk}
\usepackage[hyperref]{acl2020}
\usepackage{times}
\usepackage{graphicx}
\usepackage{latexsym}
\usepackage{booktabs} % For formal tables
\usepackage{subcaption}
\usepackage{multirow}

\usepackage{microtype}
\aclfinalcopy % Uncomment this line for the final submission
\title{Event Argument Extraction using Causal Knowledge Structures}
\author[1]{\textbf{Debanjana Kar}}
\author[2]{\textbf{Sudeshna Sarkar}}
\author[3]{\textbf{Pawan Goyal}}
\affil[ ]{Department of Computer Science \& Engineering}
\affil[ ]{Indian Institute of Technology, Kharagpur.}
\affil[1]{debanjana.kar@iitkgp.ac.in}
\affil[2,3]{\textit {\{sudeshna, pawang\}@cse.iitkgp.ac.in}}
 
\begin{document}
\maketitle
\begin{abstract}
Event Argument extraction refers to the task of extracting structured information from unstructured text for a particular event of interest. The existing works 
%model this task at a sentence level, restricting the context to a local scope. While it may be effective for short spans of text, for longer bodies of text such as news articles, it has often been observed that the arguments for an event do not necessarily occur in the same sentence as that containing an event trigger.  To tackle the issue of argument scattering across sentences, the use of global context becomes imperative in this task. Furthermore, the existing works 
exhibit poor capabilities to extract  causal event arguments like Reason and After Effects. Furthermore, most of the existing works model this task at a sentence level, restricting the context to a local scope. While it may be effective for short spans of text, for longer bodies of text such as news articles, it has often been observed that the arguments for an event do not necessarily occur in the same sentence as that containing an event trigger. To tackle the issue of argument scattering across sentences, the use of global context becomes imperative in this task. In our work, we propose an external knowledge aided approach to infuse document level event information to aid the extraction of complex event arguments.  We develop a causal network for our event-annotated dataset by  extracting relevant event causal structures from ConceptNet and phrases from Wikipedia. We use the extracted event causal features in a bi-directional transformer encoder to effectively capture long-range inter-sentence dependencies. We report the effectiveness of our proposed approach through both qualitative and quantitative analysis. In this task, we establish our findings on an event annotated dataset in 5 Indian languages. This dataset adds further complexity to the task by labeling arguments of entity type (like Time, Place) as well as more complex argument types (like Reason, After-Effect). Our approach achieves state-of-the-art performance across all the five languages. Since our work does not rely on any language specific features, it can be easily extended to other languages as well.

\end{abstract}

\section{Introduction}
\label{intro}
\input{intro}

\section{Related Work}

\input{related_work}

\section{Dataset}
We have curated an event argument annotated dataset for English and four  Indian languages, namely Bengali, Hindi, Marathi and Tamil as part of a collaborative effort. The dataset comprises of documents with sentence-level annotations of events and argument mentions.
%in the TO format. 
The data set caters specifically to the disaster domain and covers 32 event types at a fine grain level and 12 event types at a coarse level. The dataset contains annotations for 14 argument types, but in this work we focus on 6 main argument types which are, \textit{Time, Place, Casualties, After-Effect, Reason and Participant.} Almost all possible man-made and natural disaster event types have been covered in this corpus. Typically used datasets in this task like ACE 2005 and TAC KBP cater to generic events with only entity type arguments. This corpus contains arguments of both entity type and non entity type with widely varying argument boundaries. Non-entity type arguments constitute of complex argument types like \textit{Reason} and \textit{After-Effects} which may constitute one/many entities within it's span. This makes the task even more challenging and contributes to it's uniqueness.
The raw data for the corpus was collected from the FIRE document repository as well as by crawling popular news websites for the respective languages. The dataset was annotated by eight linguistic experts over a span of two years. The multi-rater Kappa agreement ratio of the annotators has been evaluated as 0.85 approximately. The data distribution is reported in Table \ref{Table-dataset} %The event-wise data distribution is as illustrated in Figure 2.
\begin{table}
\centering
\begin{tabular}{|c|c|c|c|} 
\hline
Language & Train & Valid & Test  \\ 
\hline
English  & 456   & 56    & 131   \\
Bengali  & 699   & 100   & 199   \\
Hindi    & 678   & 150   & 194   \\
Marathi  & 815   & 117   & 233   \\
Tamil    & 1085  & 155   & 311   \\
\hline
\end{tabular}
\caption{The number of documents for each language in their Train, Test and Validation splits.}
\label{Table-dataset}
\end{table}
\section{Approach}
\label{approach}
\input{approach}
\section{Results}
\input{results}
\section{Analysis}
\input{analysis}
\section{Conclusion}
In our work we have shown a pipeline approach to mine document level event labels and their arguments from each sentence in a document. We have reported a comparison study across different approaches based on multiple parameters, like input context and event information. In future, we want to extend this task to handle more languages in a multilingual setting. Our approaches have shown a huge improvement over the state-of-the art methods. In conclusion, through our work, we emphasize on the importance of both contextual and event information and report our thorough investigations on the same.

\section*{Acknowledgments}

The work done in this paper is an outcome of the project titled “A Platform for Cross-lingual and Multi-lingual Event Monitoring in Indian Languages”, supported by IMPRINT-1, MHRD, Govt. of India, and MeiTY, Govt. of India.

\bibliography{acl2020}
\bibliographystyle{acl_natbib}

\appendix

\end{document}

%% file: intro.tex
\begin{figure}[!h]
\begin{center}
%\fbox{\parbox{6cm}{
%This is a figure with a caption.}}
\includegraphics[scale=0.45]{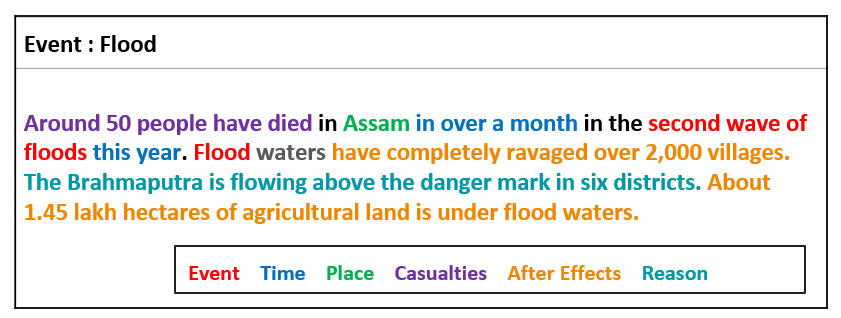} 
\caption{Sample document with annotated events and arguments. The global event category of the document is \textit{Flood} and the words in \textcolor{red}{red} indicate the event triggers.}
\label{fig.1}
\end{center}
\end{figure}
Event argument extraction is a key information extraction task that extracts structured information from unstructured texts. A widely studied task, it often involves multiple complex sub-tasks like entity retrieval, trigger detection and event-entity linking. In almost all the existing works, the terms event arguments and entities have been used interchangeably. Popular entity types which are frequently mined are that of place, time, organisations, and persons of interest. Our data set contains arguments of entity type, like Time and Place along with more complex argument types, like After Effects and Reason. These arguments with non-entity type attributes have comparatively much fewer instances in the data set than the entity type arguments. Not only that, these arguments also constitute a much more complex lexical and syntactical structure and are hard to capture using the existing architectures. In our work, we propose a novel event feature that helps capture rare complex argument instances more accurately than the existing models on this task. We construct a causal knowledge structure with the aid of external knowledge bases and for a given event, we extract its corresponding causal attribute from the structure. %Extracting arguments of both types is a superiorly challenging task, given the drastically varying boundaries different arguments may have (an argument can be a word, a phrase or an entire sentence).

%Context has always been of paramount importance for the task of event argument extraction. Prior works like \cite{nguyen-etal-2016-joint, chen2015event, subburathinam2019cross} have focused at a local scope to extract events and their attributes from unstructured text. However, as Figure  \ref{fig.1} illustrates, it is often not enough to look at individual sentences to accurately identify the arguments of an event. In the figure, the sentence  \textit{"The Brahmaputra is flowing above the danger mark in six districts."} could refer to a \textit{Reason} or an \textit{After-Effect} argument in different contexts. But on reading the document, one can understand that the document talks about a \textit{Flood} event, and that the sentence can be tagged as \textit{Reason} argument with considerable confidence. Event arguments do not necessarily co-occur in the sentence containing the event trigger. This behaviour is even more frequently observed in longer documents like news articles.  With no thematic signals obtained from event trigger words, it becomes difficult to identify the correct argument labels and their spans in such cases. In this work, we look beyond sentences for stronger contextual clues to better capture the events and their arguments from a document.\\

 Context has always been of paramount importance for the task of event argument extraction. Existing literature outlines different paradigms of modeling the contextual scope for this task. The traditional sentence level event argument extraction tasks performed on popular datasets like ACE 2005 and TAC-KBP 2017 largely restrict their contextual scope to within sentence boundaries  \cite{chen2015event, subburathinam2019cross}. However, some researchers identified the need to mine global contextual features to enhance the argument extraction capabilities and modelled the task at a multi-sentence or paragraph level \cite{yang2016joint,duan-etal-2017-exploiting, zhao-etal-2018-document}, and at a document level \cite{yang-etal-2018-dcfee, zheng2019doc2edag}. In our work, we also identify the need to explore cross-sentence contextual scope to aid accurate extraction of events and argument spans. For example as illustrated in Figure  \ref{fig.1}, the sentence  \textit{"The Brahmaputra is flowing above the danger mark in six districts."} could refer to a \textit{Reason} or an \textit{After-Effect} argument in different contexts. But on reading the document, one can understand that the document talks about a \textit{Flood} event, and that the sentence can be tagged as \textit{Reason} argument with considerable confidence. Event arguments do not necessarily co-occur in the sentence containing the event trigger. This behaviour is even more frequently observed in longer documents like news articles.  With no thematic signals obtained from event trigger words, it becomes difficult to identify the correct argument labels and their spans in such cases. In our work, we look beyond sentences for stronger contextual clues to better capture the events and their arguments from a document.
In summary, the \textbf{contributions} of our work are four fold : 
\begin{enumerate}
    \item We propose a novel event causal feature for improved extraction of complex causal arguments.
    \item Through our work, we emphasize on the effectiveness of global vs. local scope of contextual information in this task.
    \item We compare different approaches of modeling global information in this task and evaluate the effectiveness of each.
    \item We provide a novel end-to-end system for event argument extraction which beats the state-of-the art model's performance.
\end{enumerate}
\subsection{Terminology}
\label{terms}
We introduce certain terminologies to facilitate better understanding of our work .
\begin{itemize}
    \item Event : An \textbf{event} is any real life happening or occurrence which may be denoted by a word or a phrase.
    \item Event Trigger : The particular word or phrase that evokes a particular event type is known as an \textbf{event trigger}.
    \item Argument : The words or phrases which provide supporting information about the event are known as \textbf{arguments}.
    \item Causal Argument : The causal arguments refer to the \textbf{Reason} \& \textbf{After Effect} arguments of an event. The \textit{Reason} argument holds information regarding why a particular event happened whereas \textit{After Effects} argument details information about the after-math of an event. It is of importance to note here that the deaths and injuries that occur because of an event are noted under the \textit{Casualties} argument whereas all other effects of an event are covered under the \textit{After Effects} argument .
    \item Entity : Words or phrases which specifically refer to terms that represent real-world objects like people, places, organizations are known as \textbf{entities}.
\end{itemize}

%% file: related_work.tex
%\textit{"They had nothing in common but the English language."}
%- E.M. Forster

%\paragraph{Argument Extraction in Indian Languages}
Event Argument Extraction is a well researched domain, primarily in English \cite{chen2015event}, \cite{nguyen-etal-2016-joint} and Chinese \cite{lin2018nugget}, \cite{yang-etal-2018-dcfee}. While there have been efforts to extend this task to a limited number of languages like Spanish and Arabic \cite{akbik2016multilingual}, \cite{subburathinam2019cross}, the task has been hardly explored in Indian languages. In this paper, we make an effort to explore five out of the twenty two commonly spoken languages of India. The existing literature for Indian languages for this task are minimal and have either extracted arguments from unstructured text irrespective of their event links \cite{ahmadmulti}, or have jointly extracted event and argument mentions with rule based approaches \cite{patel2019tale}. Each of these works are modelled to focus at an intra-sentence local scope and do not capture inter sentence dependencies. While the sentence level extraction mechanisms outlined by prior works present a promising preface for this task in low-resource languages, the results are indicative of a huge scope of improvement - especially for arguments like \textit{Reason} which are scant on annotations.  In our work, we try to address this gap and extract sentence level events and arguments by infusing document level context. 
%Few works like \cite{zhao-etal-2018-document,zhang-etal-2020-two} have already identified that information beyond sentences can be helpful in mining events and their relevant argument spans. 
Infusing cross-sentence or global context to extract sentence level information has already shown great improvements in many related works. The work by  \cite{zhang-etal-2020-two} studies the event-argument extraction task by infusing contextual information from the five sentences surrounding an event trigger. Their work highlights the fact that learnt architectures usually capture arguments from sentences containing event triggers better than from non-event-trigger sentences. Addressing this gap, we adopt a trigger-less event detection approach in our work, similar to \cite{zheng2019doc2edag} and use document-level event information to mitigate the dependence on local event clues obtained from event triggers. The method adopted by \cite{wadden2019entity} uses a graphical span enumeration approach with a document input to model cross-task and inter-sentence dependencies. We model our work to include inter-sentential context to extract event-argument spans from input documents with greater accuracy.

%% file: approach.tex
\begin{figure}
\fbox{\includegraphics[scale=0.3]{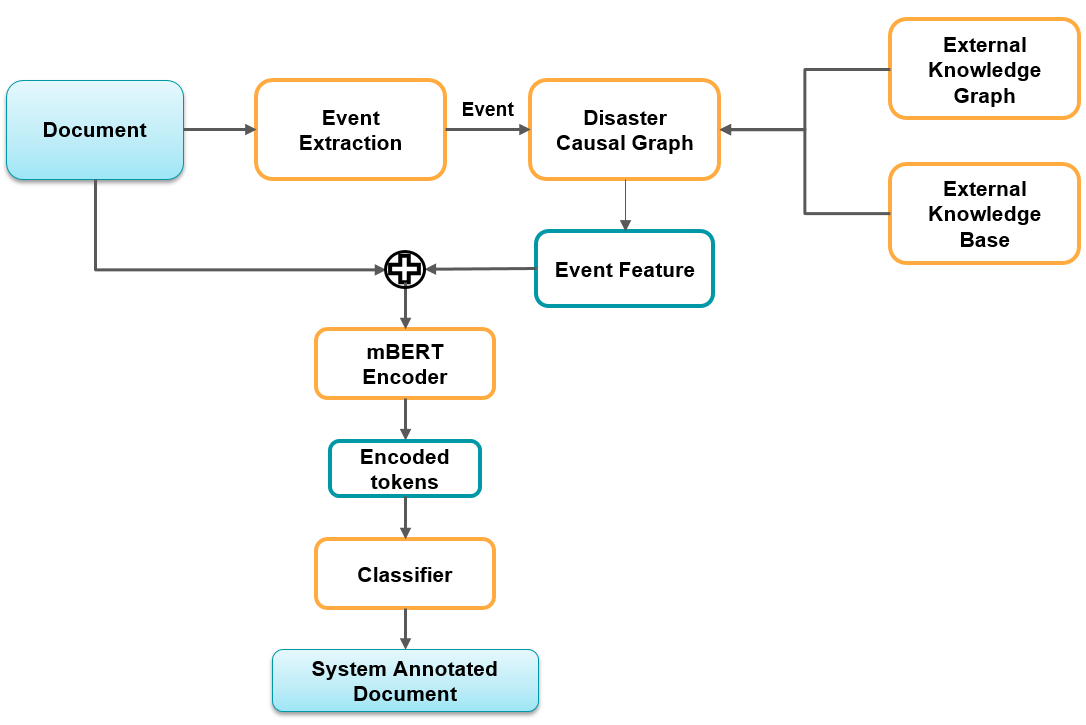}}
\caption{Overview of the general architecture.}
\label{fig-1}
\end{figure}

In this section we elaborate on the approaches we have taken to accomplish the task of extracting the event labels and it's corresponding arguments from a document. This is essentially a two-fold task: i) Document Level Event Extraction, and ii) Argument Extraction. The generic overview of our approach is illustrated in Figure \ref{fig-1}.

\subsection{Problem Formulation}
In our work, we treat the task of event-argument extraction as a sequence labeling task. Given $N$ training instances $D = {(x_{n}, y_{n})}^{N}_{n=1}$, where $x$ is a sequence of tokens from an input document and $y$ is the corresponding event-argument annotation sequence for $x$, our aim is to build a neural model which maximises the conditional probability $p_{\theta} = (y|x)$, parameterized by model parameters $\theta$. To capture events and arguments better, we propose modeling context at a global scope for this task. Infusion of global context in our work is done in two ways. First, instead of processing each sentence as a training instance, we process an entire document with sentence demarcations. Second, we infuse the document level event information into the model to aid more accurate identification of complex argument types. In the sections to follow, we have described the different approaches we have adopted to process event information into the model. We have conducted our experiments on five Indian languages. 

\subsection{Document Level Event Extraction}
Traditional approaches for event extraction involve extracting event triggers at a sentence level. We adopt a no-trigger approach to label the documents with their corresponding events but instead of adopting a distant supervision method like \cite{zheng2019doc2edag}, we adopt a supervised approach with the aid of pre-trained language models in this task. Our dataset comprises of news articles, and most often, it has been observed that the central theme of such documents are divulged in the title of the document itself. Moreover, we find that only 6\% of the total corpus contains documents with multi-event instances. Hence, we formulate this task as a multi-class sentence classification task where each document corresponds to one event category. We consider the event labels of the first event trigger instance in each document as the document's manually assigned event label.  We fine-tune a pre-trained multilingual BERT (mBERT) encoder with a linear classification head on top to classify each document title instance to an event category. Since the titles of the documents are not explicitly defined, we take the first two sentences as the title of the news article and use it as our training and testing instances. We use the the pooled output of the encoder as input to the linear classification head in the model. We find this approach to be highly effective with a very high percentage of documents being assigned their correct  event labels as shown in Table \ref{Table-Event}.

\begin{table}
\centering
\begin{tabular}{|c|c|} 
\hline
\textbf{Language} & \textbf{Micro F1-Score (\%)}  \\ 
\hline
English           & 97.71                         \\
Bengali           & 68.34                         \\
Hindi             & 87.63                         \\
Marathi           & 78.54                         \\
Tamil             & 84.89                         \\
\hline
\end{tabular}
\caption{Document Level Event Extraction Results}
\label{Table-Event}
\end{table}
%Adopting a trigger-less approach has several benefits: i) It saves 
\subsection{Argument Extraction}
The main focus of this sub-task is to extract the sentence level arguments given the document-level event information. 
%One of the biggest challenges unique to our data set are the lengths of the argument triggers which may vary over a considerably large range. They can be one word, a short phrase and at times, an entire sentence. 
%The arguments can be of entity type (like Time, Place) or more complex causal argument types (like Reason \& After-Effects). Existing works on this task \cite{ahmadmulti, patel2019tale} are unable to extract the causal argument types efficiently and exhibit very poor performance for such arguments. 
%The main goal of this task is to detect the span of the arguments related to an event, and label the argument spans correctly with special focus on the causal argument types. 
To aid the extraction of causal arguments, we introduce a causal feature along with the input document to the argument extraction module. In the following sections, we discuss how we build and use the causal feature for event argument extraction.
\begin{figure*}
    \centering
    \includegraphics[scale=0.5]{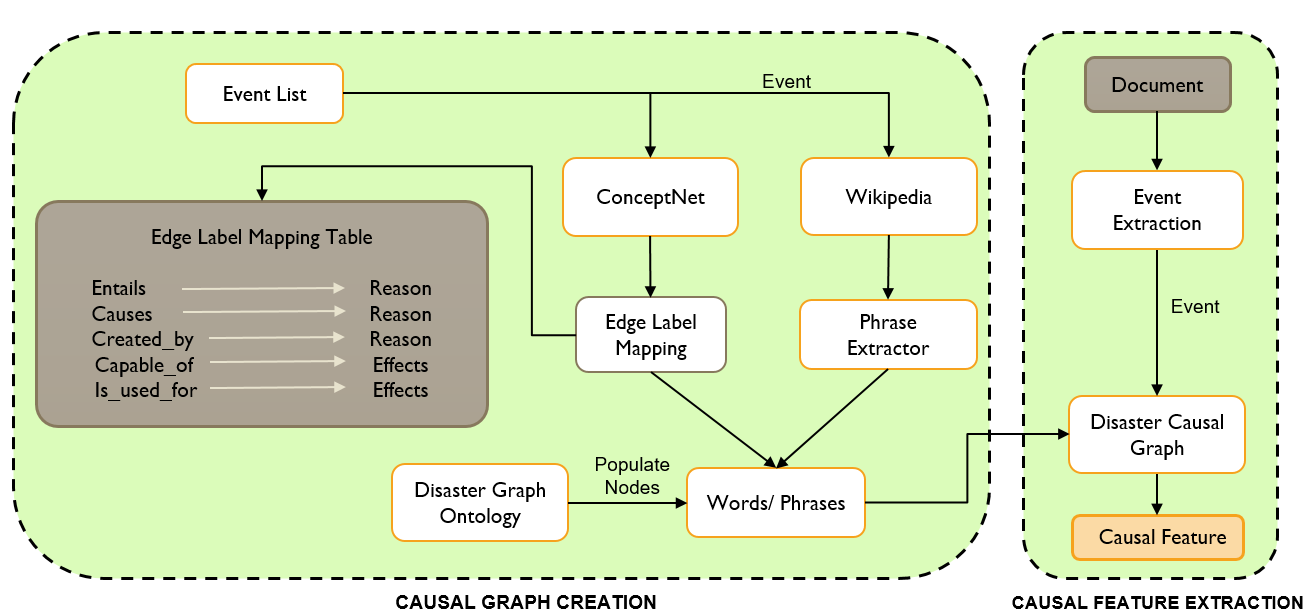}
    \caption{Detailed overview of the causal feature construction process. Since our data is of Disaster domain, we have labeled the graph in the diagram as Disaster Causal Graph.}
    \label{fig:causal_ft}
\end{figure*}
\subsubsection{Event Causal Feature} 
Unlike the generic entity type arguments like \textit{Time} and \textit{Place}, causal type arguments follow complex patterns, ranging over variable lengths and show a strong reliance on the event category. This is a natural observation - the cause and effects of an Earthquake will not be similar to that of a flood or a terrorist attack. To make the task even more challenging - these argument types, especially the \textit{Reason} argument, have very few training instances in the corpus. We define a novel event causal feature in our approach to aid the extraction of causal arguments.
\paragraph{Causal Graph Construction} We manually construct a disaster event graph ontology based on the structures defined in the annotation guidelines of the corpus as well as background knowledge of linguistic experts. Given a graph $G=(V,E)$, the vertices $V$ correspond to events and their causal argument roles (Reason \& After-Effects) and the edges $E$ show the relationships among the various events and their arguments. We manually define the event-event relationships in the graph to aid knowledge transfer across events, such that empty nodes in the graph corresponding to argument roles, can inherit knowledge for that specific argument role. 
\paragraph{Node population} The event nodes contain the event-types. The children nodes correspond to the causal argument roles. These nodes are populated using words and phrases from external knowledge bases like ConceptNet \cite{speer2017conceptnet} and Wikipedia. For ConceptNet, we have identified a few edge-relations which correspond to our causal argument labels and nodes that correspond to our event types. It is worth mentioning here, that not all the event types could be accessed in the external resources. To mitigate that, we used a list of manually curated synonyms of the event types or merged some related event types (like \textit{Transport Hazards and Vehicular Collision}) to find relevant resources in the external knowledge bases. For each event type, we query the nodes associated with the identified edges, extract the words and phrases present in the nodes and use that to populate our corresponding causal graph nodes. We also crawl the cause and effects section of relevant Wikipedia pages corresponding to the event types in our corpus. We use an unsupervised language-agnostic phrase extractor \cite{Campos2020YAKEKE} to extract phrases from the crawled Wikipedia sections and populate the corresponding causal graph nodes with the extracted phrases. Each argument node in the graph maintains a minimum of 3 and a maximum of 10 words or phrases. In the event that a certain argument node in the causal graph has less than 3 words or phrases, we exploit the defined event-event relations in the causal graph and inherit values from related event-argument nodes. The entire process is illustrated in Figure \ref{fig:causal_ft}.\\
In an additional final step, we use a sentence template to string together the words and phrases of the argument nodes of a particular event. The constructed event-causal sentence for each event will henceforth be referred to as the event causal feature in the paper.

\subsubsection{Argument Span Identification \& Classification}
Following the pipeline illustrated in Figure \ref{fig-1}, for each document, we extract it's event category and use the extracted event to obtain the corresponding event causal feature. Each input instance to the argument extraction module is augmented with the event causal feature at both ends. We train all the 12 encoder layers of the pre-trained bert-base-multilingual-cased model along with an additional linear classification head on top. The linear layer classifies the hidden states output corresponding to each token to one of the six argument types (\textit{Time, Place, Casualties, After-Effects, Reason, Participant}) or to \textit{Others}. \\
In the sections to follow, we will refer to our approach defined in this section (Section \ref{approach}) as \textbf{Event Causality Augmentation} or \textbf{ECA}.

%% file: results.tex
\begin{table*}
\centering

\begin{tabular}{|c|c|c|c|c|c|c|c|} 
\hline
\textbf{Models} & \textbf{Time}  & \textbf{Place} & \textbf{Casualties} & \textbf{After-Effects} & \textbf{Reason} & \textbf{Participant} & \textbf{Avg.}   \\ 
\hline
%DYGIE++         &                &                &                     &                        &                  &                      &                \\
Patel-Emb       & 83.05           & 88.41           & 90.09               & 55.90                  & 81.56           & 76.40                & 79.23           \\
Patel-Parallel  & 84.65          & 87.48          & 89.09               & 56.48                  & 86.31           & 75.42                & 79.90           \\
Patel-KD        & 81.74           & 87.19          & 89.83               & 47.79                  & 87.35           & 75.45                & 78.23           \\
mBERT           & 91.50          & 92.80  & 93.50               & 55.30                  & 85.90           & 84.00       & 83.80          \\

JETAE           & 90.37          & 91.84          & 92.87               & 47.42                  & 77.07            & 83.41                & 80.49           \\
EA              & 90.90          & 92.90          & 93.60               & 54.90                  & 83.60           & 86.40                & 83.71           \\
ECA (Our model)     & \textbf{92.65} & \textbf{93.41}          & \textbf{94.22}               & \textbf{58.39}                  & \textbf{90.76}           &\textbf{87.29}                   &\textbf{86.12}           \\
\hline
\end{tabular}
\caption{Argument-Wise comparison of our approach (ECA) with other baseline approaches on the English Corpus using F1-scores. The bold text refers to the best result obtained for each argument as well as overall (Avg.).}
\label{Table-overall-compare}
\end{table*}

\subsection{Experimental Settings}
We use Huggingface's bert-base-multilingual-cased model pre-trained on 104 languages\footnote{https://huggingface.co/bert-base-multilingual-cased}. Adam\cite{kingma2014adam} was used to optimize the parameters. The model was trained with a mini-batch size of 4 on a single Tesla k40-C machine and a maximum sequence length of 512. 
%Due to the small size of the datasets, the training does not consume a lot of resources. 
The models were trained for 20 epochs or until no significant changes were observed in the validation loss, whichever was earlier.
\subsection{Baselines}
We compare our work with the following enlisted approaches:
\begin{enumerate}
    \item Patel-Emb: \cite{patel2019tale} are considered the state-of-the-art results on this dataset. The works of \cite{patel2019tale} employ a set of rules to create  a rule vector for each token which embeds event class information. They use different approaches to infuse the rule vector into the model. In this approach, the rule vector was appended with each sentence token representations and fed to a single layer Bi-LSTM model.
    \item Patel-Parallel : In this approach of \cite{patel2019tale}, the rule vector and word embeddings were fed to two parallel single layer Bi-LSTM models. Their learnt hidden layer representations were concatenated to learn a joint representation of words and rules.
    \item Patel-KD: In this approach of \cite{patel2019tale}, the knowledge of rules is distilled to the Bi-LSTM network by using the rule vector to bias the weights of the neural network.
    \item mBERT: We adopt the BERT-NER from \cite{devlin-etal-2019-bert} and fine-tune the mBERT Token Classifier to extract event-argument spans and compare it with our feature based mBERT argument extractor. 
    \item JETAE: We adapt the mBERT Token Classifier for \textbf{J}oint \textbf{E}vent \textbf{T}rigger and \textbf{A}rgument \textbf{E}xtraction task as defined in \cite{patel2019tale} and report it's argument extraction capabilities as compared to our approach.
    \item EA (\textbf{E}vent \textbf{A}ugmentation): Instead of augmenting the event causal feature, we augment the event type at both ends of the document and compare it's results against our's.
\end{enumerate}
%\cite{patel2019tale} does not adjudge any of their approaches as their best model. Hence, 
We compare  the overall performance in Table \ref{Table-overall-compare} for English. For the Indian languages, we compare the F1-scores we have obtained using our approach for each argument type against the corresponding F1 score obtained from \cite{patel2019tale} in Table \ref{Table-SOTA}. Since none of the three approaches defined in \cite{patel2019tale} were reported as the best performing method for the task, for a given natural language, we have considered the highest F1-score obtained for each argument type among the three approaches as the state-of-the-art result for the same.  We report our results using precision, recall and f1-scores.
%which are based on the number of true positives, false positives and false negatives detected by the model.

\begin{table}
\centering
\begin{tabular}{|c|c|c|} 
\hline
Arg. Type     & ECA-gold & ECA-pred   \\ 
\hline
Time          & 91.84    & 92.19    \\
Place         & 71.81    & 70.82    \\
Casualties    & 82.96    & 82.79    \\
After-Effects & 53.31    & 53.13    \\
Reason        & 42.71    & 39.88    \\
Participants  & 55.11    & 54.00    \\
\hline
\end{tabular}
\caption{Results to highlight error propagation for the Bengali corpus. We report the F1-scores for each argument type with manually annotated event labels (ECA-gold) and predicted event labels (ECA-pred).}
\label{Table-error}
\end{table}

\begin{table}[!h]
\begin{tabular}{@{}c||c|c|c|c@{}}
\toprule

{\color[HTML]{333333} Lang.} & {\color[HTML]{333333} Arg. Type} & {\color[HTML]{333333} SOTA} & {\color[HTML]{333333} ECA} & {\color[HTML]{333333} $\Delta$\%} \\ \midrule
                                & Time                                 & 88.59                        & 92.19                                            & +4.06                     \\
                                & Place                                & 68.20                        & 70.82                                            & +3.84                      \\
                                & Casualties                           & 78.95                        & 82.79                                            & +4.86                      \\
                                & After Effects                        & 41.54                        & 53.13                                            & +27.90                      \\
                                & Reason                               & 17.91                        & 39.88                                            & +122.67                      \\
\multirow{-5}{*}{Bengali}       & Participant                          &49.93                        &54.00                                            & +8.15                      \\ \midrule
                                & Time                                 & 68.57                        & 72.54                                            & +5.79                     \\
                                & Place                                & 63.79                        & 73.06                                            & +14.53                     \\
                                & Casualties                           & 68.21                        & 71.93                                            & +5.45                      \\
                                & After Effects                        & 44.67                        & 45.43                                            & +1.70                      \\
                                & Reason                               & 19.25                        & 12.42                                            & -35.48                        \\
\multirow{-5}{*}{Hindi}         & Participant                          & 43.18                        & 56.24                                            & +30.25                     \\ \midrule
                                & Time                                 & 75.59                        & 80.97                                            & +7.12                      \\
                                & Place                                & 69.44                        & 74.74                                            & +7.63                      \\
                                & Casualties                           & 75.69                        & 79.48                                            & +5.01                      \\
                                & After Effects                        & 47.29                        & 59.18                                            & +25.14                     \\
                                & Reason                               & 35.93                        & 35.57                                            & -1.00                      \\
\multirow{-5}{*}{Marathi}       & Participant                          & 59.35                        & 64.55                                            & +8.76                     \\ \midrule
                                & Time                                 & 81.03                        & 89.09                                            & +9.95                     \\
                                & Place                                & 75.86                        & 83.84                                            & +10.52                    \\
                                & Casualties                           & 80.20                       & 87.21                                            & +8.74                      \\
                                & After Effects                        & 72.15                        & 83.82                                            & +16.17                     \\
                                & Reason                               & 42.72                        & 67.94                                            & +59.04                     \\
\multirow{-5}{*}{Tamil}         & Participant                          & 65.71                        & 74.61                                            & +13.54                     \\ \cmidrule(l){1-5} 
\end{tabular}
\caption{Comparison of Argument Extraction Performance of our model (ECA) with the state-of-the-art (SOTA) results (\cite{patel2019tale} for each  language. We show the $\%$ increase (+) or decrease (-) in F-Scores with respect to the SOTA results.}
\label{Table-SOTA}
\end{table}

%% file: analysis.tex
\begin{figure}
    \centering
    \includegraphics[scale=0.65]{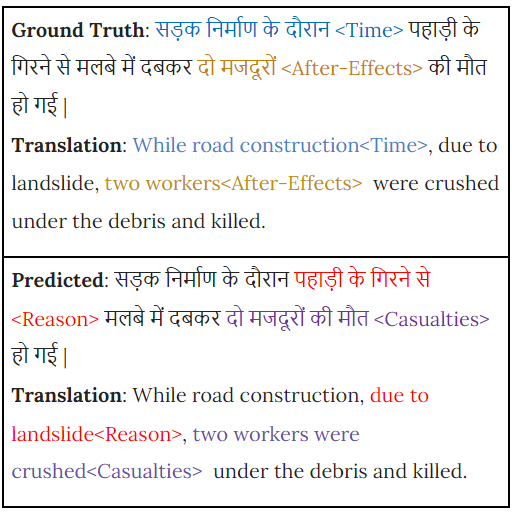}
    \caption{Manually annotated and Inferred example from the Hindi dataset highlighting the systemic confusion between \textit{event triggers} and \textit{reason} arguments. It also highlights the annotation errors existing in the corpus, which is discussed in Section \ref{case-study}.}
    \label{fig:dip-in-hindi}
\end{figure}

We perform a thorough analysis of our approaches and enlist their merits and demerits in this section. We compare our models to the state-of-the-art models established on this dataset. As reported in Table \ref{Table-overall-compare}, we establish new state-of-the-art results on this dataset. We present a comparative analysis of the approaches explored in our work and observe that our approach \textbf{E}vent \textbf{C}ausality \textbf{A}ugmentation (\textbf{ECA}) reports the best performance (Table \ref{Table-overall-compare}). In the table, we can observe a huge performance boost with the use of fine-tuned contextualised word representations of mBERT. The intended performance boost in the extraction of causal arguments is also observed on introduction of  the causal feature. We report the argument extraction performance of ECA with both manually annotated event labels (ECA-gold) and event labels predicted by our event extraction module (ECA-pred) in Table \ref{Table-error}. We specifically show these results on the Bengali dataset as it had reported the weakest event extraction results among the five languages in Table \ref{Table-Event}. In table \ref{Table-Event} we observe that the F1-scores for Bengali and Marathi are considerably low compared to the other languages. The drop in performance can be attributed to the fact that these two languages contain the maximum number of multi-event documents while our approach is based on the assumption that each document in the corpus is a single event document. Even in multi-event documents, we had observed that the dominant event was mostly present in the title of the document. The Bengali corpus consisted of a few exceptional multi-event news articles which contained news snippets about completely unrelated topics. In such cases the event extraction algorithm performs poorly, thus reporting a significant drop in performance on the Bengali dataset for the document level event extraction task. \\
We can observe the error from the event extraction module propagating to our argument extraction method in ECA-pred with the effects being visible mostly in the causal arguments in Table \ref{Table-error}. We leave the task of a joint event label identification and argument extraction as a future task, which should be able to mitigate this problem.
%For the other languages, we directly compare the state of the art results in this task on our dataset and show the \% improvement (or deterioration) for each argument in Table \ref{Table-SOTA}. It is to be noted that for almost all arguments we notice a major improvement in the performance. From this point onwards, all references to ECA in the paper are effectively made to ECA (gold). 
In both Table \ref{Table-overall-compare} and Table \ref{Table-SOTA}, we can see the desired improvement for the causal arguments. For almost all arguments we notice a major improvement in the performance in Table \ref{Table-SOTA}. In Hindi \& Marathi though, we observe a dip in the performance for the \textit{Reason} argument. We find that the dip can be attributed to two factors: i) the model confuses event triggers in certain cases as the \textit{Reason} argument - events can be the reason for other events in the document as well; ii) there are \textit{Reason} argument instances which are not annotated but gets detected by the model. We have illustrated (i) through an example in Figure \ref{fig:dip-in-hindi}. In the example, it can be observed that the model confuses the event trigger \textit{due to landslide} as the \textit{Reason} for the death of two workers. Although, it is correct in it's essence, as per the task, the phrase \textit{due to landslide} is an event trigger and not part of a \textit{Reason} argument. 
\begin{figure*}[!h]
\centering
\begin{subfigure}{.5\textwidth}
  \centering
  \includegraphics[scale=0.5]{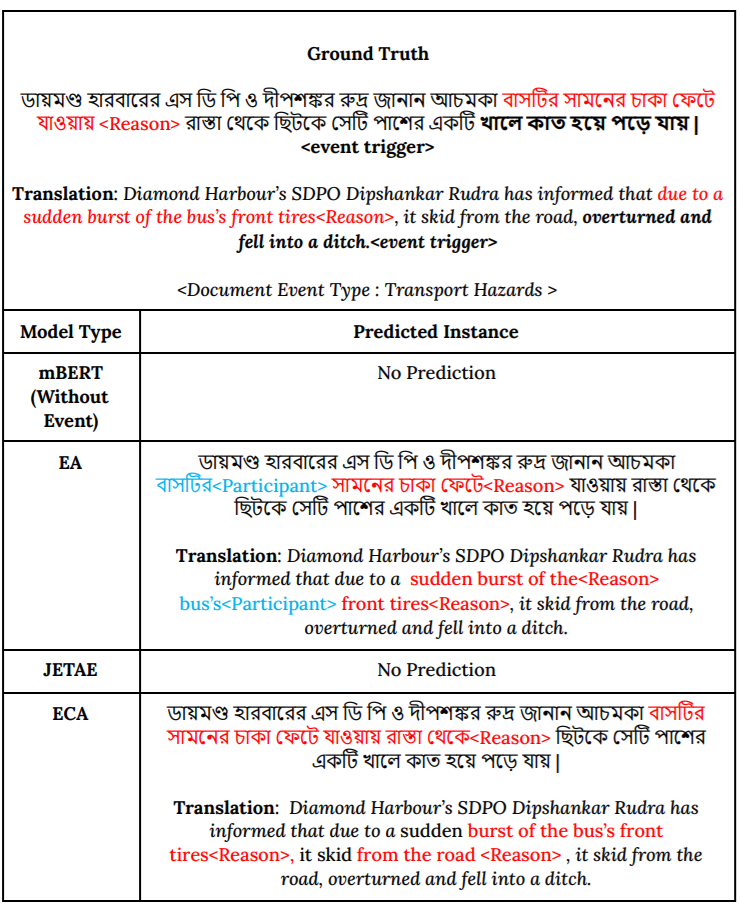}
  \caption{Example 1}
  \label{fig:sub1}
\end{subfigure}%
\begin{subfigure}{.5\textwidth}
  \centering
  \includegraphics[scale=0.5]{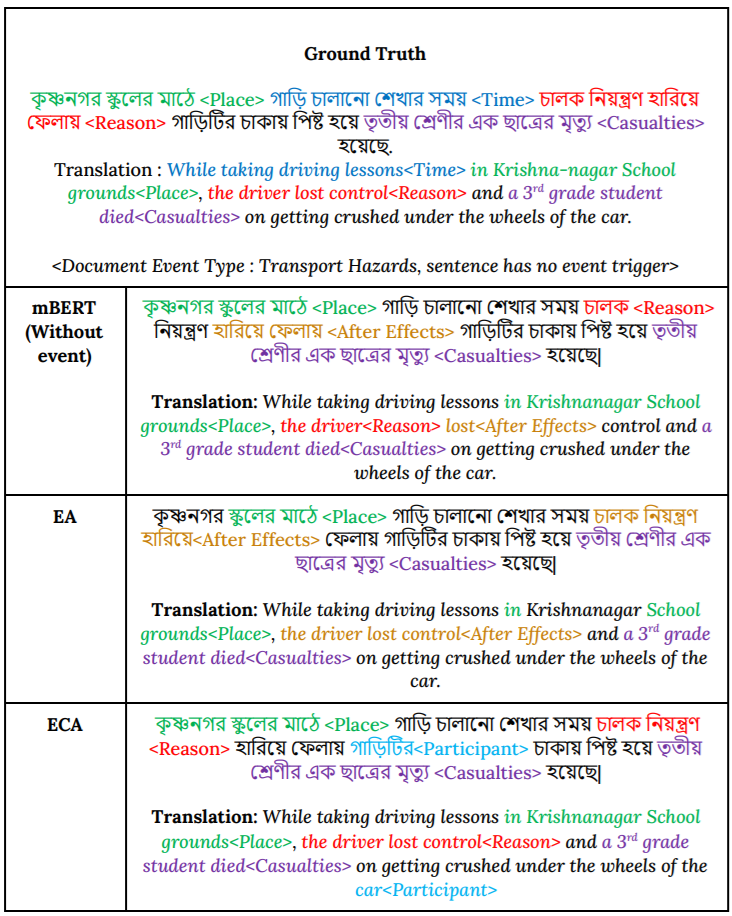}
  \caption{Example 2}
  \label{fig:sub2}
\end{subfigure}
\caption{Various inferred examples collected for a thorough qualitative analysis. The different approaches compared are our model (ECA), Event Augmentation (EA), without event (mBERT), and Joint Event Trigger and Argument Extraction model (JETAE).}
\label{fig:case_study}
\end{figure*}

\subsection{Importance of event causal information}
To investigate the importance of event information in our model for the task of argument extraction, we present our findings in Table \ref{Table-ablate}, where we compare mBERT based argument extraction models without event information (without-event), with event type information (EA) and finally with event causal feature (ECA - our model). We can see a significant improvement in performance with event information, especially with the causal feature. This rise can mainly be attributed to the significant improvement in causal argument extraction capabilities which rely heavily on event information. We also qualitatively analyse our findings in the first example cited in Figure \ref{fig:case_study}. As can be observed, without event information, it is unable to detect the presence of a \textit{Reason} Argument. With event type information, the model is able to detect the argument's presence, but we observe the model suffering from ambiguity over argument types. This ambiguity often rises with complex argument types being nested among each other, as can be observed in this example. \textit{Bus} is a Participant in this event but in this example, it is part of longer \textit{Reason} argument span. With addition of the causal feature, we find causal contexts getting learnt better and as observed in this example, causal argument spans get captured efficiently.

\subsection{Importance of document level event information}
Extracting document level event information has multiple benefits compared to event trigger extraction: i) It helps capture the overall thematic preface of the document, ii) By not extracting event triggers, we avoid the problem of ambiguous event triggers for multiple event types. For example, the word \textit{explosion} can be an event trigger for event types \textit{Terrorist Attacks, Volcano}, as well as \textit{Industrial Accidents}.; iii) It saves annotation labour and cost. Labeling document-level events is much simpler a task than annotating event triggers in each sentence of the document. By taking into account the thematic preface of the document, we will now observe it's effects on the task of event argument extraction using the same example as we used above (first example of Figure \ref{fig:case_study}). In JETAE, we often observe the model exhibiting bias towards sentences with event triggers when it comes to the task of argument extraction. In this example, since the model was unable to detect the event trigger in the sentence, it is unable to detect it's corresponding argument as well. Because of the use of document level event context, we can observe that both EA and our model, ECA are able to detect the \textit{Reason} argument span.
\subsection{Sentence vs. Paragraph vs. Document}
We study the importance of document context in the task of extracting arguments of an event. To investigate the importance of the contextual scope, we have experimented in three different settings by formatting the input instance to either include a single sentence, a paragraph or the entire document. Since paragraph boundaries were not mentioned in the dataset, we have assumed the paragraphs in our work to constitute four sentences. This decision was taken after finding the average length of the documents to be 15 sentences. We present our findings in Table \ref{Table-sent-para-doc} where we can observe a consistent increase in the F1-scores as we widen our contextual scope. The scores in each setting were evaluated by averaging the individual argument scores. While the performances at sentence and paragraph level are comparable, it is in document level context that a significant increase is observed. This increase can be mainly attributed to the effective extraction of the scattered arguments with no localised event clues to aid in their extraction process.
\subsection{Case Study}
\label{case-study}
In this section, we analyse a few inferred examples across different approaches adopted in our task. Apart from the points already discussed, we observe a few more challenges and their solutions along with their merits and demerits through our approaches. If we analyse the second example in Figure \ref{fig:case_study}, we find a classic confusion between the \textit{Reason \& After Effect} arguments. Reason and after-effects are often interchangeable concepts depending on the contextual premise. As can be observed, addition of causal feature helps resolve this ambiguity. We also observe that the model efficiently captures arguments which were otherwise missed by human annotators like the Participant argument \textit{car} in this example. This is primarily because of the document context and other similar Participant instances being annotated in likewise document context that the model is able to capture the missed annotations in the document as well. We also observe, across examples that the argument boundary mismatch persists in our approaches as well. The argument window may not be precise and may differ by a few tokens on either ends of the span.\\
Another observation that we make is more in lieu of the annotation errors in the corpus which the model has been observed to overcome in many situations. 
A confusion among annotators was observed about the semantic scope of \textit{Casualties} and \textit{After Effect} arguments as observed through the example in Figure \ref{fig:dip-in-hindi}. However, the model in most cases extracts the argument with the correct label (as also illustrated in Figure \ref{fig:dip-in-hindi}) and thus, exhibits the model's robustness against noise.
\begin{table}
\centering
\begin{tabular}{c||ccc} 
\hline
                    & P & R & F  \\ 
\hline
Without-event   &66.32   &62.57   &63.45    \\
EA    &67.21   &63.45   &64.14    \\
ECA &67.91   &65.69   &66.20    \\
%No-fine-tuning      &   &   &    \\
\hline
\end{tabular}
\caption{Comparison of different approaches on the Bengali Dataset to establish the importance of event information and event causal feature in our task.}
\label{Table-ablate}
\end{table}

\begin{table}
\centering
\begin{tabular}{c|c|c|c} 
\hline
Model & Sentence & Paragraph & Document  \\ 
\hline
EA    &59.59          &62.59           &64.14           \\
ECA &61.20          &62.42           &66.20           \\
\hline
\end{tabular}
\caption{Comparison of argument-extraction F1-scores with respect to the contextual scope, for the Bengali dataset of our corpus.}
\label{Table-sent-para-doc}
\end{table}
%\subsection{Importance of event information}